\documentclass[12]{article}
\usepackage{times}
\usepackage[numbers]{natbib}
\usepackage{amsmath,amsfonts,amssymb,times,graphicx,algorithm,algorithmic,hyperref}
\usepackage{breakurl}

\usepackage{csquotes}
\usepackage{framed}

\newcounter{lastnote}

\title{European Union regulations on algorithmic decision-making \\ and a ``right to explanation''}

\author
{Bryce Goodman,$^{1\ast}$ Seth Flaxman,$^{2}$\\
\\
\normalsize{$^{1}$Oxford Internet Institute, Oxford}\\
\normalsize{1 St Giles', Oxford OX1 3LB, United Kingdom}\\
\normalsize{$^{2}$Department of Statistics, University of Oxford,}\\
\normalsize{24-29 St Giles', Oxford OX1 3LB, United Kingdom}\\
\\
\normalsize{$^\ast$To whom correspondence should be addressed; E-mail:  flaxman@stats.ox.ac.uk.}
}
\date{}

\begin{document}
\maketitle
\begin{abstract}
We summarize the potential impact that the European Union's new General Data Protection Regulation
will have on the routine use of machine learning algorithms. Slated to take effect as law across the EU in 2018,
it will restrict automated individual decision-making (that is, algorithms that make decisions based on user-level
predictors) which ``significantly affect'' users. The law will also effectively create a ``right to explanation,'' whereby a user
can ask for an explanation of an algorithmic decision that was made about them. We argue that while this law
will pose large challenges for industry, it highlights opportunities for computer scientists to take the lead in
designing algorithms and evaluation frameworks which avoid discrimination and enable explanation.
\end{abstract}

\section{Introduction}
In April 2016, for the first time in over two decades, the European Parliament adopted a set of comprehensive regulations for the collection, storage and use of personal information, the General Data Protection Regulation (GDPR)\footnote{Regulation (EU) 2016/679 on the protection of natural persons with regard to the processing of personal data and on the free movement of such data, and repealing Directive 95/46/EC (General Data Protection Regulation) [2016] OJ L119/1.} \cite{EU_2016}.
The new regulation has been described as a ``Copernican Revolution'' in data protection law, ``seeking to shift its focus away from paper-based, bureaucratic requirements and towards compliance in practice, harmonization of the law, and individual empowerment'' \cite{Kuner_2012}. Much of the regulations are clearly aimed at perceived gaps and inconsistencies in the EU's current approach to data protection. This includes, for example, the codification of the ``right to be forgotten'' (Article 17), and regulations for foreign companies collecting data from European citizens (Article 44).

However, while the bulk of language deals with how data is collected and stored, the regulation contains
{\em Article 22: Automated individual decision-making, including profiling} (see figure \ref{fig:eu-article}) potentially prohibiting a wide swath of algorithms currently in use in, e.g.~recommendation systems, credit and insurance risk assessments, computational advertising, and social networks. This raises important issues that are of particular concern to the machine learning community. In its current form, the GDPR's requirements could require a complete overhaul of standard and widely used algorithmic techniques. The GDPR's policy on the right of citizens to receive an explanation for algorithmic decisions highlights the pressing importance of human interpretability in algorithm design. If, as expected, the GDPR takes effect in its current form in mid-2018, there will be a pressing need for effective algorithms which can operate within this new legal framework.
\begin{figure}[ht!]
\begin{framed}
\footnotesize
\centering
Article 22. Automated individual decision making, including profiling
\begin{enumerate}
\item The data subject shall have the right not to be subject to a decision based solely on automated processing, including profiling, which produces legal effects concerning him or her or similarly significantly affects him or her.
\item Paragraph 1 shall not apply if the decision:
\begin{enumerate}
\item is necessary for entering into, or performance of, a contract between the data subject and a data controller;
\item is authorised by Union or Member State law to which the controller is subject and which also lays down suitable measures to safeguard the data subject's rights and freedoms and legitimate interests; or
\item is based on the data subject's explicit consent.
\end{enumerate}
\item In the cases referred to in points (a) and (c) of paragraph 2, the data controller shall implement suitable measures to safeguard the data subject's rights and freedoms and legitimate interests, at least the right to obtain human intervention on the part of the controller, to express his or her point of view and to contest the decision.
\item Decisions referred to in paragraph 2 shall not be based on special categories of personal data referred to in Article 9(1), unless point (a) or (g) of Article 9(2) apply and suitable measures to safeguard the data subject's rights and freedoms and legitimate interests are in place.
\end{enumerate}
\end{framed}
\caption{Excerpt from the General Data Protection Regulation, \protect{\cite{EU_2016}}}
\label{fig:eu-article}
\end{figure}
\section{Background}

The General Data Protection Regulation is slated to go into effect in April 2018, and will replace the EU's 1995 Data Protection Directive (DPD). On the surface, the GDPR merely reaffirms the DPD's right to explanation and restrictions on automated decision-making. However, this reading ignores a number of critical differences between the two pieces of legislation. 

First, it is important to note the difference between a Directive and a Regulation. While a Directive ``set[s] out general rules to be transferred into national law by each country as they deem appropriate'', a Regulation is ``similar to a national law with the difference that it is applicable in all EU countries'' \cite{EU_Commission_2016}. In other words, the 1995 Directive was subject to national interpretation, and was only ever indirectly implemented through subsequent laws passed within individual member states \cite{Fromholz_2000}. The GDPR, on the other hand, requires no enabling legislation to take effect. It does not direct the law of EU member states, it simply is the law for member states (or will be, when it takes effect).

Second, the DPD and GDPR are worlds apart in terms of the penalties that can be imposed on violators. Under the DPD, there are no explicit maximum fines. Instead, fines are determined on a country by country basis. By contrast, the GDPR introduces EU-wide maximum penalties of 20 million euro or $4\%$ of global revenue, \emph{whichever is greater} (Article 83, Paragraph 5). For companies like Google and Facebook, this could mean fines in the billions. 

Third, the scope of the GDPR is explicitly global (cf. Article 3, Paragraph 1). Its requirements do not just apply to companies that are headquartered in the EU but, rather, to any companies processing EU residents' personal data. For the purposes of determining jurisdiction, it is irrelevant whether that data is processed within the EU territory, or abroad.

Before proceeding with analysis, we summarize some of the key terms employed in the GDPR as defined in {\em Article 4: Definitions}:

\begin{itemize}
\item {\bf Personal data} is ``any information relating to an identified or identifiable natural person''
\item {\bf Data subject} is the natural person to whom data relates
\item {\bf Processing} is ``any operation or set of operations which is performed on personal data or on sets of personal data, whether or not by automated means''
\item {\bf Profiling} is ``any form of automated processing of personal data consisting of the use of personal data to evaluate certain personal aspects relating to a natural person''
\end{itemize}

 Thus profiling should be construed as a subset of processing, under two conditions:
the processing is automated, and the processing is for the purposes of evaluation.

The GDPR calls particular attention to profiling aimed at ``analys[ing] or predict[ing] aspects concerning that natural person's performance at work, economic situation, health, personal preferences, interests, reliability, behavior, location or movements'' (Article 4, Paragraph 4). Given the breadth of categories, it stands to reason that the GDPR's desideratum for profiling errs on the side of inclusion, to say the least.

{\em Article 22: Automated individual decision-making, including profiling}, paragraph 1  (see figure \ref{fig:eu-article}) prohibits any ``decision based solely on automated processing, including profiling'' which ``significantly affects'' a data subject. Paragraph 2 specifies that exceptions can be made ``if necessary for entering into, or performance of, a contract'', authorized by ``Union or Member State law'' or ``based on the data subject’s explicit consent.'' However, paragraph 3 states that, even in the case of exceptions, data controllers must ``provide appropriate safeguards'' including ``the right to obtain human intervention...to express his or her point of view and to contest the decision.''
. Paragraph 4 specifically prohibits automated processing ``based on special categories of personal data'' unless ``suitable measures to safeguard the data subject’s rights and freedoms and legitimate interests are in place''.

Note that this section does not address the conditions under which it is ethically permissible to access sensitive data---this is dealt with elsewhere (e.g.~Article 7). Rather, it is implicitly assumed in this section that the data is legitimately obtained. Thus the provisions for algorithmic profiling are an additional constraint that apply even if the data processor has informed consent from data subjects.\footnote{Cf. ``consent of the data subject should not provide in itself a legal ground for processing such sensitive data'' \cite{EU_2016}}

These provisions present a number of practical challenges for the design and deployment of machine learning algorithms. This paper focuses on two: issues raised by the GDPR's stance on discrimination and the GDPR's ``right to explanation.'' Throughout, we highlight opportunities for researchers.

\section{Non-discrimination}
In general, discrimination might be defined as the unfair treatment of an individual because of his or her membership in a particular group, e.g.~race, gender, etc.~\cite{Altman_2015}. The right to non-discrimination is deeply embedded in the normative framework that underlies the EU, and can be found in Article 21 of the Charter of Fundamental Rights of the European Union, Article 14 of the European Convention on Human Rights, and in Articles 18-25 of the Treaty on the Functioning of the European Union.

The use of algorithmic profiling for the allocation of resources is, in a certain sense, inherently discriminatory: profiling takes place when data subjects are grouped in categories according to various variables, and decisions are made on the basis of subjects falling within so-defined groups. It is thus not surprising that concerns over discrimination have begun to take root in discussions over the ethics of big data. Barocas and Selbst sum the problem up succinctly: ``Big data claims to be neutral. It isn't'' \cite{Barocas_Selbst_2016}. As the authors point out, machine learning depends upon data that has been collected from society, and to the extent that society contains inequality, exclusion or other traces of discrimination, so too will the data. Consequently, ``unthinking reliance on data mining can deny members of vulnerable groups full participation in society'' \cite{Barocas_Selbst_2016}. Indeed, machine learning can reify existing patterns of discrimination---if they are found in the training dataset, then by design an accurate classifier will reproduce them. In this way, biased decisions are presented as the outcome of an 'objective' algorithm.

Paragraph 71 of the recitals (the preamble to the GDPR, which explains the rationale behind it but is not itself law) explicitly requires data controllers to ``implement appropriate technical and organizational measures'' that ``prevents, inter alia, discriminatory effects'' on the basis of processing sensitive data. According to {\em Article 9: Processing of special categories of personal data}, sensitive data includes:

\begin{displayquote}
personal data revealing racial or ethnic origin, political opinions, religious or philosophical beliefs, or trade-union membership, and the processing of genetic data, biometric data for the purpose of uniquely identifying a natural person, data concerning health or data concerning a natural person's sex life or sexual orientation...
\end{displayquote}

It is important to note that paragraph 71 and Article 22 paragraph 4 specifically address discrimination from profiling that makes use of sensitive data. In unpacking this mandate, we must distinguish between two potential interpretations. The first {\em minimal interpretation} is that this directive only pertains to cases where an algorithm is making direct use of data that is explicitly sensitive. This would include, for example, variables that code for race, finances, or any of the other categories of sensitive information referred to in Article 9. However, it is widely acknowledged that simply removing certain variables from a model does not ensure predictions that are, in effect, uncorrelated to those variables (e.g.~\cite{Leese_2014,Hardt_2014}). For example, if a certain geographic region has a high number of low income or minority residents, an algorithm that employs geographic data to determine loan eligibility is likely to produce results that are, in effect, informed by race and income.

Thus a second {\em maximal  interpretation}, takes a broader view of `sensitive data' to include not only those variables which are explicitly named, but also any variables with which they are correlated. This would put the onus on a data processor to ensure that algorithms are not provided with datasets containing variables that are correlated with the ``special categories of personal data'' in Article 9.

However, this interpretation also suffers from a number of complications in practice. With relatively small datasets it may be possible to both identify and account for correlations between sensitive and `non-sensitive' variables. However,  removing all data correlated with sensitive variables may make the resulting predictor virtually useless. As Calders and Verwer note, ''postal code can reveal racial information and yet at the same time, still give useful, non-discriminatory information on loan defaulting'' \cite{Calders_Verwer_2010}.

Furthermore, as datasets become increasingly large, correlations can become increasingly complex and difficult to detect. The link between geography and income may be obvious, but less obvious correlations---say between IP address and race---are likely to exist within large enough datasets and could lead to discriminatory effects. 
For example, at an annual conference of actuaries, consultants from Deloitte explained that they can now ``use thousands of `non-traditional' third party data sources, such as consumer buying history, to predict a life insurance applicant's health status with an accuracy comparable to a medical exam'' \cite{Robinson_Yu_Rieke_2014}. With sufficiently large data sets, the task of exhaustively identifying and excluding data features correlated with ``sensitive categories'' {\em a priori} may be impossible. Companies may also be reluctant to exclude certain covariates---web-browsing patterns are a very good predictor for various recommendation systems, but they are also correlated with sensitive categories.

A final challenge, which purging variables from the dataset does not address, is posed by what we term \emph{uncertainty bias}. This bias arises when two conditions are met:
\begin{itemize}
\item{One group is underrepresented in the sample\footnote{Note that the underrepresentation of a minority in a sample can arise through historical discrimination or less access to technology, but it is also a feature of a random sample in which groups are by construction represented at their population rates. In public health and public policy research, minorities are sometimes oversampled to address this problem.}, so there is more uncertainty associated with predictions about that group}
\item{The algorithm is \emph{risk averse}, so it will {\em ceteris paribus} prefer to make decisions based on predictions about which they are more confident (i.e.~those with smaller confidence intervals \citep{Aigner_Cain_1977})}
\end{itemize}

\begin{figure}[ht!]
    \centering
    \includegraphics[width=.8\paperwidth]{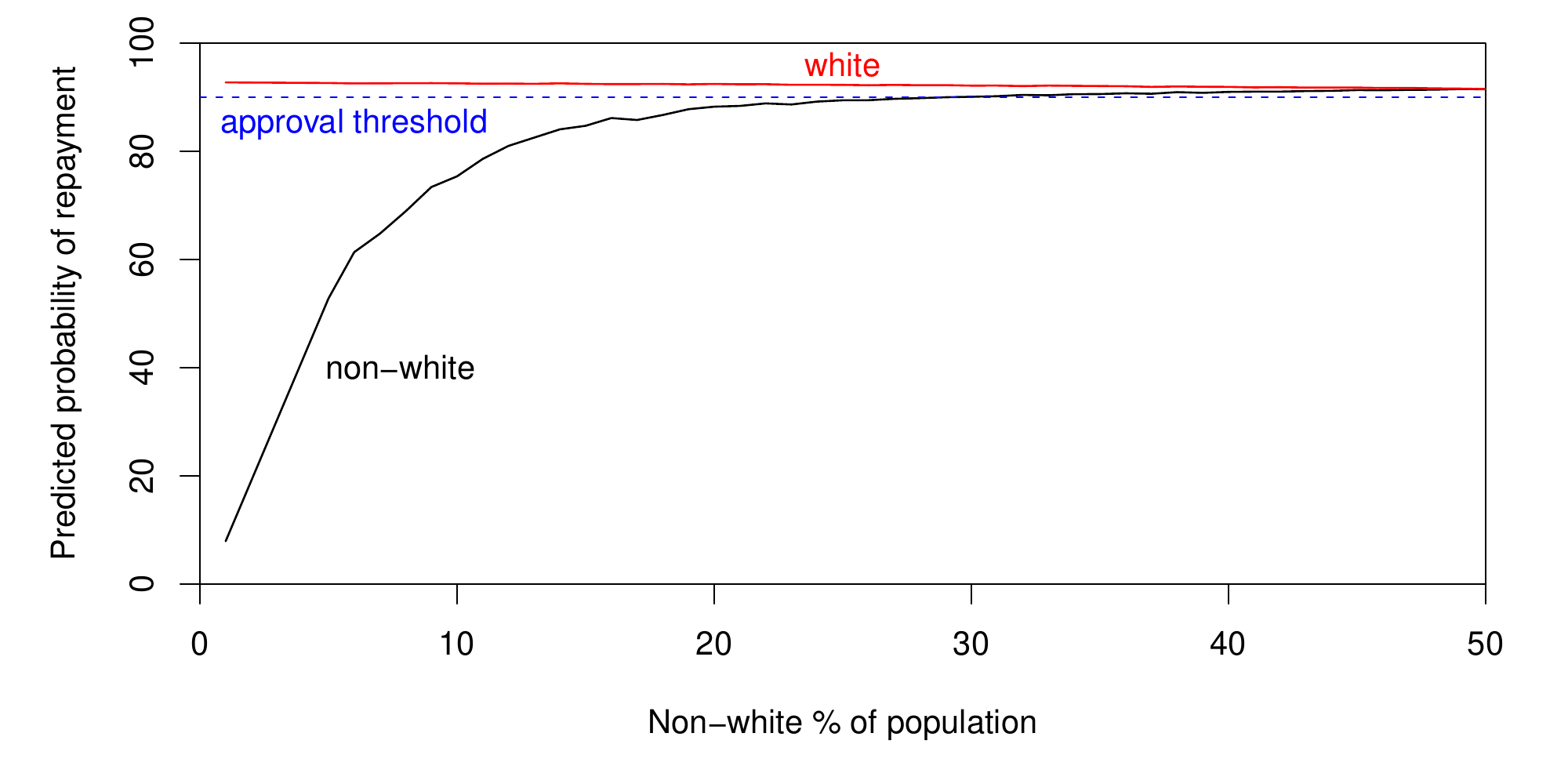}
    \caption{An illustration of uncertainty bias: a hypothetical algorithm is used to predict the probability of loan repayment in a setting in which the ground truth is that non-whites and whites are equally likely to repay. The algorithm is risk averse, so it makes an offer when the lower end of the 95\% confidence interval for its predictions lie above a fixed approval threshold of 90\% (dashed line). When non-whites are less than 30\% of the population, and assuming a simple random sample, the algorithm exhibits what we term ``uncertainty bias''---the underrepresentation of non-whites means that predictions for non-whites have less certainty, so they are not offered loans. As the non-white percentage approaches 50\% the uncertainty approaches that of whites and everyone is offered loans.}
    \label{fig:uncertainty-bias}
\end{figure}

In practice, this could mean that predictive algorithms (e.g.~for loan approval) favor groups that are better represented in the training data, since there will be less uncertainty associated with those predictions.
Uncertainty bias is illustrated in Figure \ref{fig:uncertainty-bias}. The population consists of two groups, white and non-whites. 
An algorithm is used to decide whether to extend a loan, based on the predicted probability that the individual will
repay the loan. We repeatedly generated synthetic datasets of size 500, varying the true proportion of non-whites in the population. 
In every case, we set the true probability of repayment to be independent of group membership: all individuals have a 95\% probability of repayment regardless of race. Using a logistic regression classifier, we consider a case in which loan decisions are made in a risk averse manner, by using the following decision rule: check whether the lower end of the 95\% confidence interval for an individual is above a fixed ``approval threshold'' of 90\%. In all cases, all white individuals will be offered credit since the true probability is 95\% and the sample size is large enough for the confidence interval to be small. However, when the non-white population is any fraction less than 30\% of the total population, they will not be extended credit due to the uncertainty inherent in the small sample.

Note that in practice, more complicated combinations of categories (occupation, location, consumption patterns, etc.) would be considered by a classifier and rare combinations will have very few observations.
This issue is compounded in an active learning setting: consider the same setting, where non-whites and whites are equally likely to default. A small initial bias towards the better represented groups due will be compounded over time as the active learning acquires more examples of the better represented group and their overrepresentation grows.
The GDPR thus presents us with a dilemma with two horns: under the minimal interpretation the non-discrimination requirement is ineffective, under the maximal interpretation it is infeasible. However it would be premature to conclude that non-discrimination measures are without merit. Rather, the complexity and multifaceted nature of algorithmic discrimination suggests that appropriate solutions will require an understanding of how it arises in practice. This highlights the need for human-intelligible explanations of algorithmic decision making.

\section{Right to explanation}
The provisions outlined in Articles 13-15 specify that data subjects have the right to access information collected about them, and also requires data processors to ensure data subjects are notified about the data collected. However, it is important to distinguish between these rights, which may be termed the right to access and notification, and additional ``safeguards for the rights and freedoms of the data subject'' required under Article 22 when profiling takes place. Although the Article does not elaborate what these safeguards are beyond ``the right to obtain human intervention''\footnote{The exact meaning and nature of the intended intervention is unspecified, and the requirement raises a number of important questions that are beyond our current scope.}, Articles 13 and 14 state that, when profiling takes place, a data subject has the right to ``meaningful information about the logic involved.''  This requirement prompts the question: what does it mean, and what is required, to explain an algorithm's decision?

Standard supervised machine learning algorithms for regression or classification are inherently based on discovering reliable associations / correlations to aid in accurate out-of-sample prediction, with no concern for causal reasoning or ``explanation'' beyond the statistical sense in which it is possible to measure the amount of variance explained by a predictor. As Mildebrandt writes, ``correlations stand for a probability that things will turn out the same in the future. What they do not reveal is why this should be the case'' \cite{Hildebrandt_2008}. The use of algorithmic decisions in an increasingly wider range of applications has led some (e.g.~\cite{Pasquale_2015}) to caution against the rise of a ``black box'' society and demand increased transparency in algorithmic decision-making. The nature of this requirement, however, is not always clear.

Burrell distinguishes between three barriers to transparency \cite{Burrell_2016}:

\begin{itemize}
\item Intentional concealment on the part of corporations or other institutions, where decision making procedures are kept from public scrutiny
\item Gaps in technical literacy which mean that, for most people, simply having access to underlying code is insufficient
\item A ``mismatch between the mathematical optimization in high-dimensionality characteristic of machine learning and the demands of human-scale reasoning and styles of interpretation''
\end{itemize}

Within the GDPR, {\em Article 13: Information to be made available or given to the data subject} goes some way\footnote{It is not clear whether companies will be required to disclose their learning algorithms or training datasets and, if so, whether that information will be made public.} towards the first barrier, stipulating that data processors inform data subjects when and why data is collected, processed, etc. {\em Article 12: Communication and modalities for exercising the rights of the data subject} attempts to solve the second by requiring that communication with data subjects is in ``concise, intelligible and easily accessible form.'' The third barrier, however, poses additional challenges that are particularly relevant to algorithmic selection and design. As Lisboa  notes, ``machine learning approaches are alone in the spectrum in their lack of interpretability'' \cite{Lisboa_2013}.

Putting aside any barriers arising from technical fluency, and also ignoring the importance of training the model,
it stands to reason that an algorithm can only be explained if the trained model
can be articulated and understood by a human. It is reasonable to suppose that any adequate explanation would, at a minimum, provide an account of how input features relate to predictions, allowing one to answer questions such as:
Is the model more or less likely to recommend a loan if the applicant is a minority?
Which features play the largest role in prediction?

There is of course a tradeoff between the representational capacity of a model and its interpretability, ranging
from linear models (which can only represent simple relationships but are easy to interpret)
to nonparametric methods like support vector machines and Gaussian processes
(which can represent a rich class of functions but are hard to interpret). 
Ensemble methods like random forests pose a particular challenge, as predictions result from an aggregation or averaging
procedure. Neural networks, especially with the rise of deep learning, pose perhaps the biggest challenge---what hope
is there of explaining the weights learned in a multilayer neural net with a complex architecture?
These issues have recently gained attention within the machine learning community and are becoming an active area of research \citep{WHI2016}.
One promising avenue of research concerns developing algorithms to quantify the degree of influence of input variables on outputs, given black-box access to a trained prediction algorithm \citep{dattaalgorithmic2016}. 

\section{Conclusion}

This paper has focused on two sets of issues raised by the forthcoming GDPR that are directly relevant to machine learning: the right to non-discrimination and the right to explanation. This is by no means a comprehensive overview of the potential challenges that will be faced by engineers as they adapt to the new framework. The ability of humans to intervene in algorithmic decision making, or for data subjects to provide input to the decision-making process, will also likely impose requirements on algorithmic design and require further investigation.

While the GDPR presents a number of problems for current applications in machine learning they are, we believe, good problems to have. The challenges described in this paper emphasize the importance of work that ensures that algorithms are not merely efficient, but transparent and fair.  Research is underway in pursuit of rendering algorithms more amenable to ex post and ex ante inspection \cite{dattaalgorithmic2016,Vellido_Martin-Guerrero_Lisboa_2012,jia2016recombination}. Furthermore, a number of recent studies have attempted to tackle the issue of discrimination within algorithms by introducing tools to both identify \cite{Berendt_Preibusch_2012,Sandvig_Hamilton_Karahalios_Langbort_2014} and rectify \cite{Calders_Verwer_2010,Hajian_Domingo-Ferrer_Martinez-Balleste_2011,Zliobaite_Kamiran_Calders_2011,Berendt_Preibusch_2014,Dive_Khedkar_2014, Feldman2015} cases of unwanted bias. It remains to be seen whether these techniques are adopted in practice. One silver lining of this research is to show that, for certain types of algorithmic profiling, it is possible to both identify and implement interventions to correct for discrimination. This is in contrast to cases where discrimination  arises from human judgment. The role of extraneous and ethically inappropriate factors in human decision making is well documented (e.g., \cite{Tversky_Kahneman_1974,Danziger_Levav_Avnaim-Pesso_2011,Abrams_Bertrand_Mullainathan_2012}), and discriminatory decision making is pervasive in many of the sectors where algorithmic profiling might be introduced (e.g.~\cite{Holmes_Horvitz_1994,Bowen_Bok_1998}). We believe that, properly applied, algorithms can not only make more accurate predictions, but offer increased transparency and fairness over their human counterparts (cf. \cite{Laqueur_Copus_2015}).

Above all else, the GDPR is a vital acknowledgement that, when algorithms are deployed in society, few if any decisions are purely ``technical''. Rather, the ethical design of algorithms requires coordination between technical and philosophical resources of the highest caliber. A start has been made, but there is far to go. And, with less than two years until the GDPR takes effect, the clock is ticking.


\newpage
\bibliographystyle{apa}
\bibliography{euregs2}

\begin{thebibliography}{}

\bibitem[\protect\astroncite{Abrams
  et~al.}{2012}]{Abrams_Bertrand_Mullainathan_2012}
Abrams, D., Bertrand, M., and Mullainathan, S. (2012).
\newblock Do judges vary in their treatment of race?
\newblock {\em Journal of Legal Studies}, 41(2):347--383.

\bibitem[\protect\astroncite{Aigner and Cain}{1977}]{Aigner_Cain_1977}
Aigner, D.~J. and Cain, G.~G. (1977).
\newblock Statistical theories of discrimination in labor markets.
\newblock {\em Industrial and Labor Relations Review}, 30(2):175.

\bibitem[\protect\astroncite{Altman}{2015}]{Altman_2015}
Altman, A. (2015).
\newblock Discrimination.
\newblock In Zalta, E.~N., editor, {\em The Stanford Encyclopedia of
  Philosophy}. Stanford University, fall 2015 edition.

\bibitem[\protect\astroncite{Barocas and Selbst}{2016}]{Barocas_Selbst_2016}
Barocas, S. and Selbst, A.~D. (2016).
\newblock Big data's disparate impact.
\newblock {\em California Law Review}, 104.

\bibitem[\protect\astroncite{Berendt and
  Preibusch}{2012}]{Berendt_Preibusch_2012}
Berendt, B. and Preibusch, S. (2012).
\newblock {\em Exploring Discrimination: A User-centric Evaluation of
  Discrimination-Aware Data Mining}, pages 344--351.

\bibitem[\protect\astroncite{Berendt and
  Preibusch}{2014}]{Berendt_Preibusch_2014}
Berendt, B. and Preibusch, S. (2014).
\newblock Better decision support through exploratory discrimination-aware data
  mining: foundations and empirical evidence.
\newblock {\em Artificial Intelligence and Law}, 22(2):175--209.

\bibitem[\protect\astroncite{Bowen and Bok}{1998}]{Bowen_Bok_1998}
Bowen, W.~G. and Bok, D. (1998).
\newblock {\em The Shape of the River. Long-Term Consequences of Considering
  Race in College and University Admissions.}
\newblock ERIC.

\bibitem[\protect\astroncite{Burrell}{2016}]{Burrell_2016}
Burrell, J. (2016).
\newblock How the machine ``thinks'': Understanding opacity in machine learning
  algorithms.
\newblock {\em Big Data \& Society}, 3(1).

\bibitem[\protect\astroncite{Calders and Verwer}{2010}]{Calders_Verwer_2010}
Calders, T. and Verwer, S. (2010).
\newblock Three naive bayes approaches for discrimination-free classification.
\newblock {\em Data Mining and Knowledge Discovery}, 21(2):277--292.

\bibitem[\protect\astroncite{Danziger
  et~al.}{2011}]{Danziger_Levav_Avnaim-Pesso_2011}
Danziger, S., Levav, J., and Avnaim-Pesso, L. (2011).
\newblock Extraneous factors in judicial decisions.
\newblock {\em Proceedings of the National Academy of Sciences},
  108(17):6889--6892.

\bibitem[\protect\astroncite{Datta et~al.}{2016}]{dattaalgorithmic2016}
Datta, A., Sen, S., and Zick, Y. (2016).
\newblock Algorithmic transparency via quantitative input influence.
\newblock {\em 37th IEEE Symposium on Security and Privacy}.

\bibitem[\protect\astroncite{Dive and Khedkar}{2014}]{Dive_Khedkar_2014}
Dive, R. and Khedkar, A. (2014).
\newblock An approach for discrimination prevention in data mining.
\newblock {\em International Journal of Application or Innovation in
  Engineering and Management}, 3(6).

\bibitem[\protect\astroncite{{European Commission}}{2016}]{EU_Commission_2016}
{European Commission} (2016).
\newblock Legislation.

\bibitem[\protect\astroncite{Feldman et~al.}{2015}]{Feldman2015}
Feldman, M., Friedler, S.~A., Moeller, J., Scheidegger, C., and
  Venkatasubramanian, S. (2015).
\newblock {\em Certifying and removing disparate impact}, pages 259--268.
\newblock ACM.

\bibitem[\protect\astroncite{Fromholz}{2000}]{Fromholz_2000}
Fromholz, J.~M. (2000).
\newblock {The European Union Data Privacy Directive}.
\newblock {\em Berkeley Technology Law Journal}, 15(1):461--484.

\bibitem[\protect\astroncite{Hajian
  et~al.}{2011}]{Hajian_Domingo-Ferrer_Martinez-Balleste_2011}
Hajian, S., Domingo-Ferrer, J., and Martinez-Balleste, A. (2011).
\newblock {\em Discrimination prevention in data mining for intrusion and crime
  detection}, page 47�54.
\newblock IEEE.

\bibitem[\protect\astroncite{Hardt}{2014}]{Hardt_2014}
Hardt, M. (2014).
\newblock How big data is unfair: Understanding sources of unfairness in data
  driven decision making.

\bibitem[\protect\astroncite{Hildebrandt}{2008}]{Hildebrandt_2008}
Hildebrandt, M. (2008).
\newblock {\em Defining profiling: a new type of knowledge?}, page 17�45.
\newblock Springer.

\bibitem[\protect\astroncite{Holmes and Horvitz}{1994}]{Holmes_Horvitz_1994}
Holmes, A. and Horvitz, P. (1994).
\newblock Mortgage redlining: Race, risk, and demand.
\newblock {\em Journal of Finance}, page 81�99.

\bibitem[\protect\astroncite{Jia and Liang}{2016}]{jia2016recombination}
Jia, R. and Liang, P. (2016).
\newblock Data recombination for neural semantic parsing.
\newblock In {\em Association for Computational Linguistics (ACL)}.

\bibitem[\protect\astroncite{{Kim} et~al.}{2016}]{WHI2016}
{Kim}, B., {Malioutov}, D.~M., and {Varshney}, K.~R. (2016).
\newblock {Proceedings of the 2016 ICML Workshop on Human Interpretability in
  Machine Learning (WHI 2016)}.
\newblock {\em ArXiv e-prints}.

\bibitem[\protect\astroncite{Kuner}{2012}]{Kuner_2012}
Kuner, C. (2012).
\newblock {The European Commission's} proposed data protection regulation: A
  {Copernican revolution in European data protection law}.
\newblock {\em Bloomberg BNA Privacy and Security}, 6(2012):1�15.

\bibitem[\protect\astroncite{Laqueur and Copus}{2015}]{Laqueur_Copus_2015}
Laqueur, H. and Copus, R. (2015).
\newblock Machines learning justice: The case for judgmental bootstrapping of
  legal decisions.
\newblock {\em Available at SSRN}.

\bibitem[\protect\astroncite{Leese}{2014}]{Leese_2014}
Leese, M. (2014).
\newblock The new profiling: Algorithms, black boxes, and the failure of
  anti-discriminatory safeguards in the european union.
\newblock {\em Security Dialogue}, 45(5):494�511.

\bibitem[\protect\astroncite{Lisboa}{2013}]{Lisboa_2013}
Lisboa, P.~J. (2013).
\newblock {\em Interpretability in Machine Learning�Principles and Practice},
  page 15�21.
\newblock Springer.

\bibitem[\protect\astroncite{{Parliament and Council of the European
  Union}}{2016}]{EU_2016}
{Parliament and Council of the European Union} (2016).
\newblock {General Data Protection Regulation}.

\bibitem[\protect\astroncite{Pasquale}{2015}]{Pasquale_2015}
Pasquale, F. (2015).
\newblock {\em The Black Box Society: The Secret Algorithms That Control Money
  and Information}.
\newblock Harvard University Press, 1 edition edition.

\bibitem[\protect\astroncite{Robinson et~al.}{2014}]{Robinson_Yu_Rieke_2014}
Robinson, D., Yu, H., and Rieke, A. (2014).
\newblock {\em Civil Rights, Big Data, and Our Algorithmic Future}.
\newblock Social Justice and Technology.

\bibitem[\protect\astroncite{Sandvig
  et~al.}{2014}]{Sandvig_Hamilton_Karahalios_Langbort_2014}
Sandvig, C., Hamilton, K., Karahalios, K., and Langbort, C. (2014).
\newblock Auditing algorithms: Research methods for detecting discrimination on
  internet platforms.
\newblock {\em Data and Discrimination: Converting Critical Concerns into
  Productive Inquiry}.

\bibitem[\protect\astroncite{Tversky and
  Kahneman}{1974}]{Tversky_Kahneman_1974}
Tversky, A. and Kahneman, D. (1974).
\newblock Judgment under uncertainty: Heuristics and biases.
\newblock {\em science}, 185(4157):1124�1131.

\bibitem[\protect\astroncite{Vellido
  et~al.}{2012}]{Vellido_Martin-Guerrero_Lisboa_2012}
Vellido, A., Mart�n-Guerrero, J.~D., and Lisboa, P.~J. (2012).
\newblock {\em Making machine learning models interpretable.}, volume~12, pages
  163--172.
\newblock Citeseer.

\bibitem[\protect\astroncite{Zliobaite
  et~al.}{2011}]{Zliobaite_Kamiran_Calders_2011}
Zliobaite, I., Kamiran, F., and Calders, T. (2011).
\newblock {\em Handling Conditional Discrimination}, pages 992--1001.

\end{thebibliography}

\end{document}